\documentclass[pdflatex,sn-nature]{sn-jnl}
\usepackage{graphicx}
\usepackage{multirow}
\usepackage{amsmath,amssymb,amsfonts}
\usepackage{amsthm}
\usepackage{mathrsfs}
\usepackage[title]{appendix}
\usepackage{xcolor}
\usepackage{textcomp}
\usepackage{manyfoot}
\usepackage{booktabs}
\usepackage[utf8]{inputenc}
\usepackage[T1]{fontenc}
\usepackage{subcaption}
\usepackage{siunitx}
\usepackage{pdfpages}

\newcommand{\Vc}{V_{\mathrm c}}
\newcommand{\Km}{K_{\mathrm m}}

\newcommand{\had}{\odot}

\begin{document}

\title{Learning pharmacokinetic dynamics with structured neural ODEs for precision dosing}

\author*[1]{\fnm{Benjamin} \sur{Maurel}}\email{benjamin.maurel@inserm.fr}
\author[1]{\fnm{Agathe} \sur{Guilloux}}
\author[1]{\fnm{Sarah} \sur{Zohar}}
\author[1]{\fnm{Moreno} \sur{Ursino}}
\equalcont{Equal contribution}
\author[2,3]{\fnm{Jean-Baptiste} \sur{Woillard}}
\equalcont{Equal contribution}

\affil[1]{\orgdiv{HeKA}, \orgname{Université Paris Cité, Inria, Inserm}, \city{Paris}, \postcode{F-75015}, \country{France}}
\affil[2]{\orgdiv{P\&T, U1248}, \orgname{Université de Limoges, CHU Limoges}, \city{Limoges}, \country{France}}
\affil[3]{\orgdiv{Service de Pharmacologie, Toxicologie et Pharmacovigilance}, \orgname{CHU Limoges}, \city{Limoges}, \country{France}}

\abstract{Individualized dosing requires predicting how a patient's drug exposure would change under a different dose or schedule. However, a model that accurately predicts exposure at the current dose may not reliably predict what happens when treatment changes. We developed two neural models that learn pharmacokinetic dynamics from data without predefined pharmacokinetic equations. The first predicts exposure after a dose change, while the second also allows changes in dosing schedules. In simulations, the second model predicted exposure under new dosing schedules more accurately than competing neural models. On real data of kidney transplant recipients treated with tacrolimus, both models showed lower prediction errors than Bayesian estimation when predicting exposure at a later visit. These findings suggest that incorporating pharmacokinetic principles into neural models can improve predictions needed for individualized dosing.}
\keywords{Model-informed precision dosing, Tacrolimus, Neural ordinary differential equations, Counterfactual prediction, Therapeutic drug monitoring, Kidney transplantation}

\maketitle

\section*{Introduction}

\begin{figure}[tbp]
\centering
\includegraphics[width=0.95\textwidth]{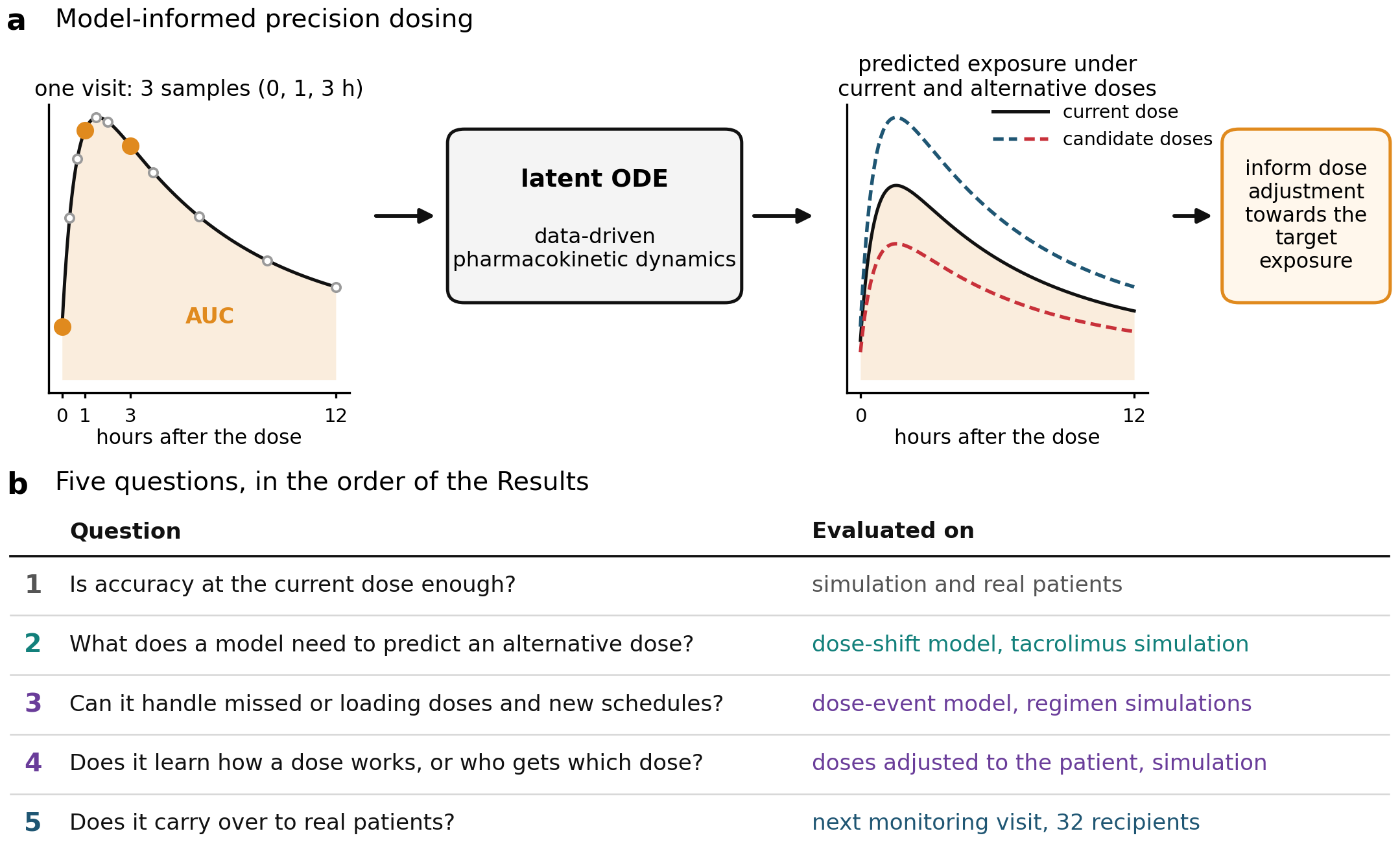}
\caption{\textbf{Study overview.} \textbf{a}, Role of the models in model-informed precision dosing. From the samples collected under the current dose (orange), the models predict the exposure (AUC) under the observed regimen and under alternative doses or regimens. \textbf{b}, The five questions of the study, in the order of the Results.}
\label{fig:task}
\end{figure}

Tacrolimus is the backbone of immunosuppression after kidney transplantation. Its therapeutic index is narrow and its pharmacokinetics vary widely between patients, requiring individualised dose adjustments guided by therapeutic drug monitoring (TDM)~\cite{brunet2019therapeutic}. The area under the concentration--time curve over a dosing interval (AUC) is estimated from two or three blood samples, and the dose is adjusted towards a target exposure~\cite{Saint-Marcoux_2013} (Fig.~\ref{fig:task}a). When the estimated exposure falls outside the therapeutic target, a dose adjustment may be required. This decision needs a prediction for a dose the patient has not received. Given the samples collected under the current dose, what exposure would the same patient achieve under an alternative dose or schedule? Such a prediction is called counterfactual. The standard approach to this question relies on population pharmacokinetic models, typically formulated as nonlinear mixed-effects models, whose individual parameters are estimated from the observed samples by maximum a posteriori Bayesian estimation (MAP-BE) and subsequently used to predict exposure under alternative dosing regimens~\cite{sheiner1982bayesian,davidian2017nonlinear,woillard2011population}. However, these models require several assumptions regarding their structure, including the number of compartments, absorption and elimination processes, covariate effects, between-patient variability and residual error. Individual exposure predictions may be biased when the model does not adequately describe the underlying pharmacokinetic processes~\cite{greppmair2023towards}.

Machine learning approaches have been proposed to overcome some of these limitations. Limited-sampling regressions and gradient-boosted trees have been used to estimate tacrolimus exposure from a few blood samples and patient covariates~\cite{zhao2011limited,woillard2021tacrolimus}. However, these approaches primarily predict exposure under the observed dosing regimen, without explicitly modelling how exposure changes under alternative doses, which population models can simulate~\cite{keutzer2022machine}. Neural ordinary differential equations (ODEs) use neural networks to learn the dynamics of a system over time~\cite{chen2018neural}. They generate concentration--time profiles from these learned dynamics, without a prescribed compartmental structure, elimination model or covariate--parameter relationships. Latent ODEs~\cite{rubanova2019latent} infer the initial state of the dynamics from sparse and irregularly sampled observations, which resembles the individualisation performed by MAP-BE. Neural ODEs have been applied to pharmacokinetic data~\cite{bram2024low,giacometti2025leveraging}, to predict patient-specific responses from early observations~\cite{lu2021nmi} and to predict drug exposure under alternative dosing regimens~\cite{lu2021neural,baaz2026empirical}.

A prediction under an alternative dose can be computed as soon as the curve is generated by dynamics, but whether it is correct depends on what the model has learned about how a dose affects drug exposure. In clinical practice, each patient is observed under the dose prescribed to them. The samples provide information about exposure under this dose, but not necessarily about how exposure in the same patient would change under an alternative dose. A flexible model may therefore attribute the observed exposure to patient characteristics, the administered dose, or both, and its accuracy at the observed dose does not show which explanation it has learned. In addition, the dose and the clearance of the patient (the rate at which the drug is eliminated) are often correlated, because TDM adjusts doses towards a target exposure~\cite{ahn2005inherent}. A model may then learn this association, which reflects the dosing policy, and underestimate the pharmacokinetic effect of the dose.

A population model can answer dosing questions because of three properties of its structure, regardless of its particular equations. First, the exposure is generated by dynamics from the state of the patient, allowing exposure to be estimated from a few samples. Second, an administration increments the drug state without changing the parameters governing the dynamics, allowing exposure to be predicted for the same patient under a different dose. Third, the drug state (the amounts in the compartments) is distinct from the physiology of the patient (the parameters), so a dosing history of any length acts on the drug state while leaving patient physiology unchanged.

\begin{figure}[tbp]
\centering
\includegraphics[width=0.60\textwidth]{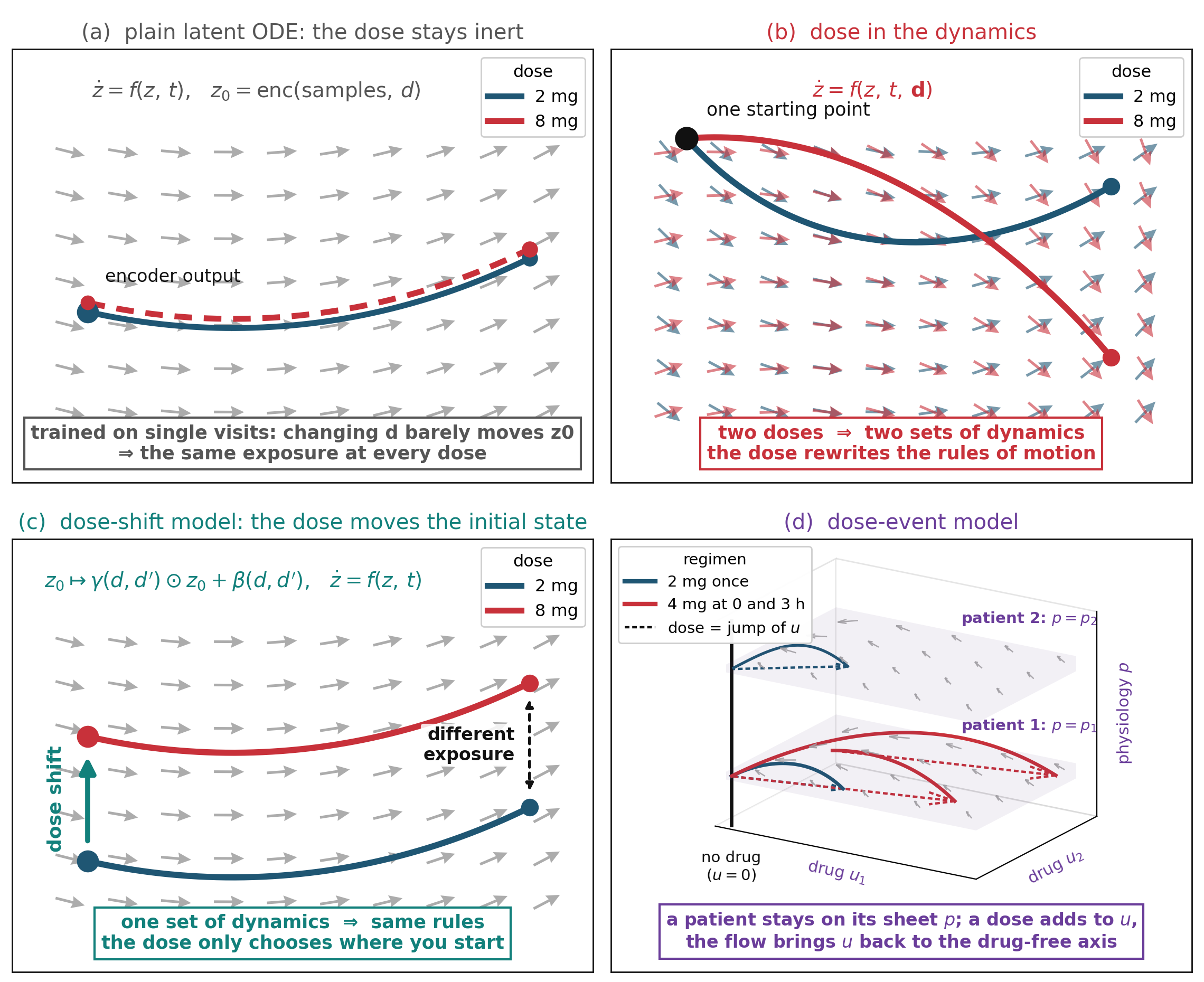}
\caption{\textbf{Where the dose acts.} Schematic latent dynamics (not model output) for two doses (\textbf{a}--\textbf{c}) or two regimens (\textbf{d}), in blue and red. \textbf{a}, Plain latent ODE. The dose enters only the encoder. \textbf{b}, Dose-in-the-dynamics model, $\dot z = f(z,t,d)$, with one set of learned dynamics per dose. \textbf{c}, Dose-shift model. The dose moves the initial state, and one set of dynamics that does not depend on the dose governs the motion. \textbf{d}, Dose-event model. Each administration translates the drug coordinates $u$ along $b(p)$, and the physiology $p$ is unchanged.}
\label{fig:designs}
\end{figure}

In this work, we incorporate these properties into latent ODEs that learn the underlying dynamics directly from data, in two models suited to different dosing questions. The dose-shift model addresses the question raised at a typical tacrolimus monitoring visit, the exposure that would follow a change of dose on the same schedule. It is trained on pairs of visits of the same patient at different doses, and the dose moves the latent state while the learned dynamics do not depend on the dose (Fig.~\ref{fig:designs}c). The dose-event model is intended for regimens with several administrations at arbitrary times, such as loading, missed or additional doses and changes of schedule. Its latent state is split into drug coordinates, which each administration increments, and physiology coordinates, which no administration changes (Fig.~\ref{fig:designs}d). We first examine whether accuracy at the observed dose shows that a model can predict an alternative dose. We then evaluate the two models in simulations, in which the exposure of every patient under any dose is known, with doses assigned at random and doses adjusted to the patient. Finally, we test them on the next monitoring visit of kidney transplant recipients (Fig.~\ref{fig:task}b).

\section*{Results}

\subsection*{Accurate prediction at the current dose does not guarantee accuracy at an alternative dose}

\begin{table}[tb]
\centering
\footnotesize
\setlength{\tabcolsep}{4pt}
\caption{\textbf{Comparison of the latent ODE and Bayesian estimation for the prediction of the observed AUC under different model specifications.} MAP-BE uses a population model with linear elimination and no haematocrit effect, correct in Scenario~1 and misspecified in Scenarios~2 and~3 (Methods). Mean $\pm$ s.d.\ over 100 simulated datasets. MPE, mean percentage error. RMSPE, root-mean-square percentage error. $^{\dagger}$From our rerun (Methods).}
\label{tab:scenarios}
\begin{tabular}{@{}llcc@{}}
\toprule
\textbf{Scenario} & \textbf{Metric} & \textbf{Latent ODE} & \textbf{MAP-BE} \\
\midrule
1: correctly specified model & MPE (\%) & $+1.23 \pm 2.10$ & $+1.06 \pm 1.66$ \\
 & RMSPE (\%) & $16.72 \pm 1.14$ & $16.63 \pm 1.23$ \\
\midrule
2: omitted covariate effect & MPE (\%) & $+0.36 \pm 2.52$ & $+2.35 \pm 1.47$ \\
 & RMSPE (\%) & $20.42 \pm 1.19$ & $22.96 \pm 0.73$ \\
\midrule
3: misspecified elimination model & MPE (\%) & $+0.39 \pm 2.46$ & $+1.02 \pm 7.49$ \\
 & RMSPE (\%) & $15.04 \pm 2.71^{\dagger}$ & $23.57 \pm 8.55$ \\
\bottomrule
\end{tabular}
\end{table}

We report the error of the predicted AUC relative to the true AUC. Table~\ref{tab:scenarios} and the clinical cohorts at the observed dose use the root-mean-square percentage error (RMSPE) and the mean percentage error (MPE). All other comparisons use the median absolute percentage error and the share of patients within $\pm20\%$ of the true AUC, the range accepted for tacrolimus~\cite{Saint-Marcoux_2013}.

A population model can give biased predictions when its equations are wrong, and a latent ODE learns its dynamics from the data. We first asked whether this gives an advantage at the observed dose, in three simulated scenarios in which the population model was correct (Scenario~1), missed the effect of haematocrit on clearance (Scenario~2), or assumed linear elimination while elimination was saturable (Scenario~3). From three samples at 0, 1 and 3~h after the dose, the plain latent ODE and Bayesian estimation (MAP-BE) had similar prediction errors with a correct population model (RMSPE 16.7\% against 16.6\%). With an omitted covariate effect and with a misspecified elimination model, the latent ODE was more precise, by 2.5 and 8.5 points of RMSPE (Table~\ref{tab:scenarios}).

In real kidney transplant recipients, the latent ODE was at least as accurate as Bayesian estimation. In repeated cross-validation on the development cohort, in which almost every test profile had other visits of the same patient in the training set, its mean RMSPE was 7.99\%, against 9.24\% for the iterative two-stage Bayesian estimator used in clinical practice (it2B$_{\text{ISBA}}$)~\cite{Saint-Marcoux_2013}. On an independent external cohort of 72 patients~\cite{marquet2018comparative}, it reached 10.45\%, against 11.58\% for this estimator and 11.48\% for MAP-BE under a population model fitted by stochastic approximation expectation--maximisation (SAEM)~\cite{kuhn2005maximum} (Supplementary Table~3 and Supplementary Fig.~1).

\begin{table}[tb]
\centering
\footnotesize
\setlength{\tabcolsep}{3pt}
\caption{\textbf{Predictive performance of neural ODEs and of Bayesian estimation under observed and alternative doses.} Simulated two-visit tacrolimus cohort, 100 training and 1100 test patients. Med., median absolute percentage error of the AUC, mean (s.d.) over 10 training runs (Lu et al., three). $\pm$20\%, share of patients within $\pm20\%$ of the true AUC. $>$100\%, number of patients with an error above 100\%. Arrows show the better direction. $^{a}$One run with a level bias (Supplementary Note~3). $^{b}$Trained on single visits, as in ref.~\cite{lu2021nmi}. $^{c}$With covariates and the cross-visit loss. $^{d}$Two of three runs (Supplementary Note~3). MAP-BE, Bayesian estimation from the three visit-1 samples (one fit). $^{e}$The simulator's structure. $^{f}$Visit-1 AUC of the first MAP-BE row multiplied by the dose ratio.}
\label{tab:benchmark}
\begin{tabular}{@{}lccccc@{}}
\toprule
 & \multicolumn{2}{c}{\textbf{Visit 1}} & \multicolumn{3}{c}{\textbf{Visit 2}} \\
 & \multicolumn{2}{c}{observed dose} & \multicolumn{3}{c}{alternative dose, 1--8 mg} \\
\cmidrule(lr){2-3}\cmidrule(lr){4-6}
\textbf{Model} & Med.\,$\downarrow$ & $\pm$20\%\,$\uparrow$ & Med.\,$\downarrow$ & $\pm$20\%\,$\uparrow$ & $>$100\%\,$\downarrow$ \\
\midrule
\multicolumn{6}{@{}l}{\textit{Latent neural ODEs, no pharmacokinetic equations}} \\
Dose-shift model$^{a}$ & 10.0 (1.2) & 80.4 & 13.2 (1.2) & 67.5 & 5.6 \\
Dose-in-the-dynamics model & 11.1 (0.8) & 75.5 & 13.9 (0.8) & 66.3 & 8.5 \\
Lu et al., original$^{b}$ & 11.9 (0.6) & 77.0 & 60.4 (3.7) & 10.3 & 249.3 \\
Lu et al., same task$^{c,d}$ & 12.7 (2.8) & 71.9 & 15.9 (0.1) & 60.0 & 7.0 \\
\midrule
\multicolumn{6}{@{}l}{\textit{MAP-BE, population model fitted by SAEM on the same 100 patients}} \\
Saturable elimination$^{e}$ & 7.8 & 90.0 & 9.9 & 79.0 & 1 \\
Linear elimination & 8.7 & 85.3 & 29.5 & 32.9 & 41 \\
Dose-proportional rule$^{f}$ & 7.8 & 90.0 & 31.4 & 31.4 & 45 \\
\bottomrule
\end{tabular}
\end{table}

Predicting the exposure that a patient received does not show that a model can predict the exposure under another dose. To test this, we simulated kidney transplant recipients observed at two visits with different doses, so that the true AUC of each patient is known at both. Every model receives the three samples of visit~1 with their dose and the covariates, and predicts the AUC at that dose and at an alternative dose between 1 and 8~mg that it is given but never observes. The two-port model of Lu et al.~\cite{lu2021nmi}, trained as proposed on single visits, predicted the observed dose well (11.9\%), but its error at the alternative dose was 60.4\%, with a response to the dose about four times too weak (Table~\ref{tab:benchmark}). The model introduced in the next section showed the same pattern when trained without comparing visits (9.8\% and 83.1\%, Supplementary Table~1). The samples of a visit already set its exposure level, so a model can fit each visit without learning the effect of the dose.

\subsection*{Trained on patients seen at two doses, a neural ODE predicts exposure at an alternative dose without pharmacokinetic equations}

We therefore built a model in which the dose acts as it does in a compartmental model, and we trained it on changes of dose. In a compartmental model, each administration adds a drug amount to the depot compartment, which modifies the drug state without altering the parameters governing its dynamics. The dose-shift model applies this principle to a latent state, a learned summary of the patient and the drug that is inferred from the samples. The dose moves this state, and the same learned dynamics, which do not depend on the dose, then generate the concentration curve (Fig.~\ref{fig:designs}c). The model is trained on pairs of visits of the same patient at different doses, so that the samples of each visit have to predict the other visit.

At the alternative dose, the dose-shift model, which is given no equations, reached a median error of 13.2\%, with 67.5\% of patients within $\pm20\%$ (Table~\ref{tab:benchmark}). Bayesian estimation with the simulator's structure, under a population model fitted on the same 100 patients, reached 9.9\% and 79.0\%, so the dose-shift model came within about three points of this reference. With linear elimination, as in the published tacrolimus model~\cite{woillard2011population}, Bayesian estimation predicted the observed dose almost as well, but its error at the alternative dose rose to 29.5\%, with 32.9\% of patients within $\pm20\%$, and the dose-proportional rule had a similar error (31.4\%). The dose-shift model therefore had less than half the prediction error of Bayesian estimation under this linear assumption, with twice as many patients within $\pm20\%$, and the accuracy at the observed dose did not show which approach would hold at another dose. On these training patients, each seen at two doses, comparing the two structures selected saturable elimination (Bayesian information criterion 10,980 against 12,260), and the selected model reached 10.9\%.

Training on two visits made the largest difference. It reduced the error at the alternative dose of the dose-shift model from 83.1\% to 16.4\% (Supplementary Table~1) and, with the covariates added, that of the two-port model of Lu et al.\ from 60.4\% to 15.9\%. Where the dose acts added about one further point. With the dose given to the learned dynamics and the same pairs of visits (dose-in-the-dynamics model, Fig.~\ref{fig:designs}b, Methods), the error was 13.9\%, against 13.2\% for the dose-shift model, which was lower by 1.2 points in a patient-by-patient comparison (95\% bootstrap interval 0.2 to 1.9) and had the fewest large errors among the neural models.

The dose-shift model also held on a more realistic change of dose, with each patient at steady state on a first dose, switched to a second and observed four days later. It reached a median error of 12.3\%, against 37.5\% for the empirical Bayes model (EB-VAE) of Baaz et al.~\cite{baaz2026empirical}, in which the dose enters the learned dynamics (Supplementary Note~2).

\subsection*{Separating drug from patient allows prediction under regimens never seen in training}

Clinical dosing decisions often involve more than a single change of dose, for example doses at arbitrary times, a missed or loading dose, or a change of schedule. The dose shift was learned only on states at the start of a dosing interval. Applied a few hours after the dose, its prediction departed further from the one made at the dose as the delay grew, and a regression predicting the central volume (a patient parameter that relates the amount of drug in the blood to its concentration) from the latent state lost all predictive value one hour after the dose (Supplementary Fig.~2). Constraining the model to conserve a block of latent coordinates did not solve this problem, because its readout left that block nearly unused (Supplementary Note~3). We therefore separated drug and physiology states, as in a compartmental model. The latent state of the dose-event model is split into drug coordinates, to which each administration adds in proportion to its amount, and physiology coordinates, which no administration changes (Fig.~\ref{fig:designs}d). The physiology coordinates stayed fixed under any regimen, so a single regression on them predicted the central volume equally well at every time.

\begin{table}[tb]
\centering
\scriptsize
\setlength{\tabcolsep}{1.2pt}
\caption{\textbf{Prediction under regimens never seen in training.} Median absolute percentage error of the AUC, mean $\pm$ s.d.\ over training runs. \textbf{a}, AUC(0--30~h). Original, the five regimens of ref.~\cite{baaz2026empirical}. New, five further regimens fixed before these runs. Titrated, the high dose given preferentially to fast eliminators. Training-regimen errors are in Supplementary Table~5. Bold, best learned model. $^{a}$Regimen model~\cite{lu2021neural}. $^{b}$Two-port model~\cite{lu2021nmi}. $^{c}$Regimens with the training schedule and equal doses (three original, two new). --, not run (Methods). \textbf{b}, AUC(0--45~h), randomised doses. Runs lower, training runs in which the dose-event model is more accurate than the empirical Bayes model. Bold, lower mean error. The models of Lu et al.\ are in Supplementary Table~4.}
\label{tab:regimens}
\begin{tabular}{@{}lcccccc@{}}
\toprule
\multicolumn{7}{@{}l}{\textbf{a}\quad Design of Baaz et al., one compartment, linear elimination} \\
\midrule
 & \multicolumn{2}{c}{\textbf{Sparse, randomised}} & \multicolumn{2}{c}{\textbf{Sparse, titrated}} & \multicolumn{2}{c}{\textbf{Dense, randomised}} \\
\cmidrule(lr){2-3}\cmidrule(lr){4-5}\cmidrule(lr){6-7}
\textbf{Model} & Original & New & Original & New & Original & New \\
\midrule
\multicolumn{7}{@{}l}{\textit{Dynamics learned from data}} \\
Empirical Bayes model & $14.4\pm1.5$ & $29.2\pm4.9$ & $32.4\pm3.6$ & $30.0\pm4.2$ & $5.2\pm2.3$ & $22.1\pm3.8$ \\
Dose-event model & $\mathbf{10.9\pm1.8}$ & $\mathbf{8.2\pm1.5}$ & $\mathbf{12.3\pm2.9}$ & $\mathbf{12.2\pm4.5}$ & $\mathbf{2.6\pm0.3}$ & $\mathbf{2.2\pm0.4}$ \\
Lu et al., regimen$^{a}$ & $40.6\pm31.2$ & -- & $42.4\pm28.9$ & -- & $40.0$ & -- \\
Lu et al., two-port$^{b}$ & $50.5\pm0.4$ & -- & $42.9\pm1.1$ & -- & $64.6$ & -- \\
\midrule
\multicolumn{7}{@{}l}{\textit{Simulator's equations, population fitted by SAEM}} \\
MAP-BE & $2.6\pm0.2$ & $2.6\pm0.2$ & $2.6\pm0.2$ & $2.6\pm0.1$ & $0.7\pm0.0$ & $0.7\pm0.0$ \\
Dose-proportional rule$^{c}$ & $2.9\pm0.1$ & $2.8\pm0.2$ & $2.8\pm0.1$ & $2.8\pm0.2$ & $1.2\pm0.1$ & $1.1\pm0.0$ \\
\bottomrule
\end{tabular}

\vspace{6pt}
\setlength{\tabcolsep}{3.5pt}
\begin{tabular}{@{}llccc@{}}
\toprule
\multicolumn{5}{@{}l}{\textbf{b}\quad Tacrolimus simulator, five regimens never seen in training} \\
\midrule
\textbf{Elimination} & \textbf{Samples} & \textbf{Dose-event model} & \textbf{Empirical Bayes model} & \textbf{Runs lower} \\
\midrule
Linear & 6 & $17.2\pm3.2$ & $\mathbf{15.9\pm5.3}$ & 3/5 \\
 & 12 & $\mathbf{16.5\pm2.2}$ & $25.5\pm6.1$ & 5/5 \\
 & 61 & $\mathbf{5.1\pm0.5}$ & $17.1\pm1.8$ & 3/3 \\
\midrule
Saturable & 6 & $17.4\pm3.2$ & $\mathbf{16.6\pm6.2}$ & 2/5 \\
 & 12 & $\mathbf{19.7\pm10.4}$ & $26.1\pm7.1$ & 4/5 \\
 & 61 & $\mathbf{12.7\pm0.1}$ & $22.4\pm3.9$ & 3/3 \\
\bottomrule
\end{tabular}
\end{table}

We first tested regimens never seen in training on the simulated design of Baaz et al.~\cite{baaz2026empirical}, which has one compartment and linear elimination, with 100 patients per arm receiving 400 or 800~mg and observed with 6 samples (sparse sampling) or 61 samples (dense sampling). The dose-event model was chosen among variants evaluated on the regimens of this design (Supplementary Table~2). Table~\ref{tab:regimens}a therefore reports later runs with the model frozen and the comparisons specified in advance, on new training sets, new test patients and five further regimens. All these comparisons held (Supplementary Note~3).

With sparse sampling, the dose-event model predicted these regimens almost as accurately as the training regimen (10.9\% and 8.2\%, against 8.3\%, Supplementary Table~5). It was more accurate than the empirical Bayes model of Baaz et al.\ in eight of nine training runs on their five regimens (14.4\%) and in every run on the further regimens (29.2\%). The difference was largest on a loading dose of 1200~mg, above every training amount (7.5\% against 91.0\%), and the dose-event model was also more accurate on the other four further regimens (8.3\% against 13.8\%). With dense sampling, it stayed within 4\% on every regimen.

Neither model of Lu et al.\ accurately predicted exposure under these regimens, with errors of 40 to 65\%. Given the simulator's equations, Bayesian estimation reached 2.6\%, and the dose-event model came closest to this reference among the models that learn the dynamics from data. Adding the dose to the drug coordinates mattered most for changes of schedule, since a dose input of the learned dynamics raised the errors on the regimens with doses at 0, 3 and 8~h (Supplementary Table~2).

\begin{figure}[tbp]
\centering
\includegraphics[width=0.85\textwidth]{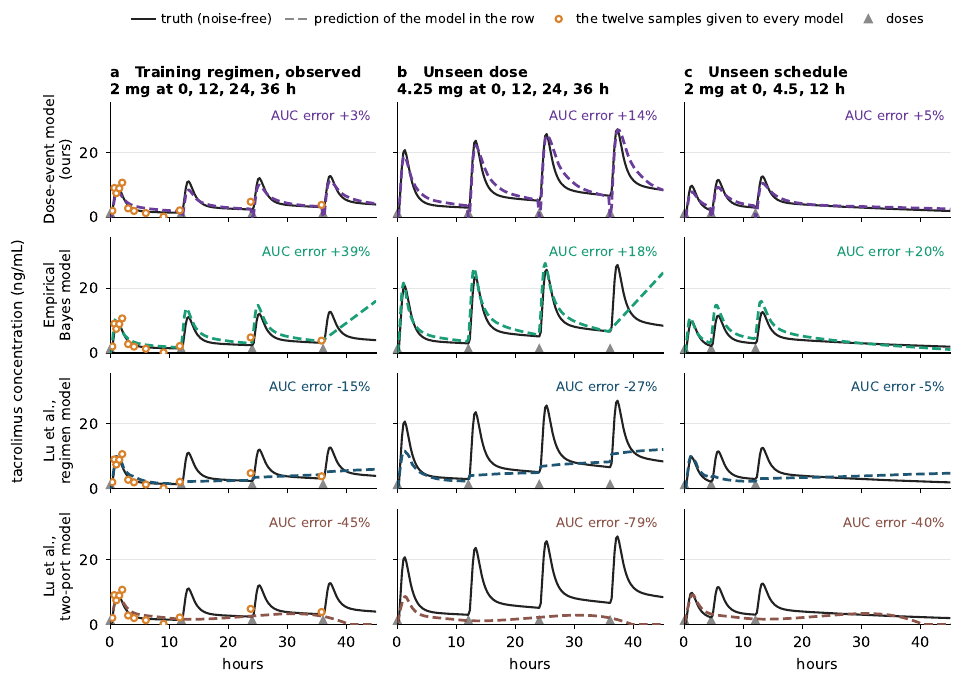}
\caption{\textbf{What the models predict for one patient.} One illustrative test patient. Tacrolimus simulator, linear elimination, 12 samples. \textbf{a}, The training regimen, under which the samples given to every model were collected. \textbf{b}, \textbf{c}, Two regimens never seen in training, predicted from the same samples. Each row is one model, with the error of its predicted AUC over 0--45~h.}
\label{fig:tacro-patients}
\end{figure}

We repeated the comparison on the tacrolimus simulator, under linear and saturable elimination, with 6, 12 or 61 samples (Table~\ref{tab:regimens}b). With six samples, the dose-event model and the empirical Bayes model had similar errors. With richer sampling, the dose-event model was more accurate, for example 16.5\% against 25.5\% with 12 samples and linear elimination, in each of five training runs. Fig.~\ref{fig:tacro-patients} shows the predictions for one test patient from the 12 samples of its training regimen. The dose-event model followed both unseen regimens. The empirical Bayes model over-predicted them and drifted upwards after the last dose, the regimen model of Lu et al.\ missed the peaks of the higher dose, and the two-port model under-predicted every regimen.

The two models answer different dosing questions. For a change of dose on the same schedule, the dose-shift model remains the more accurate (13.2\% against 15.0\% at the alternative dose), and the dose-event model covers other regimens.

\subsection*{Doses adjusted to the patient can distort what a model learns about the dose}

\begin{figure}[tbp]
\centering
\includegraphics[width=0.95\textwidth]{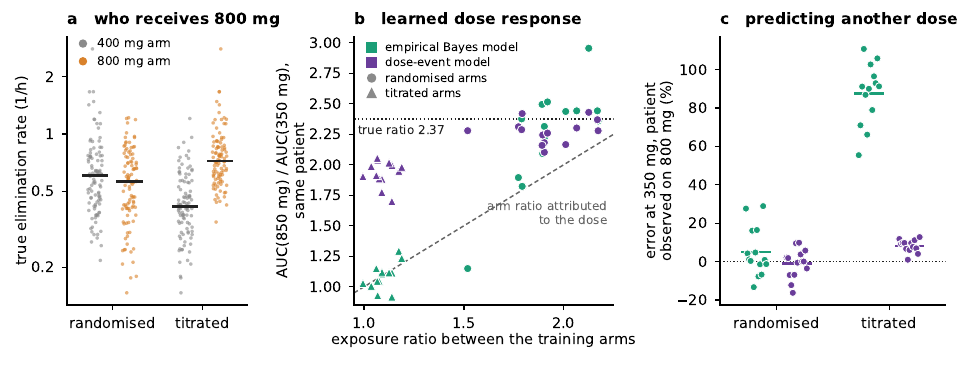}
\caption{\textbf{Effect of dose titration on learned dose--exposure relationships} (design of Baaz et al., linear simulator, sparse sampling). \textbf{a}, Elimination rate of the patients of each training arm, with randomised and with titrated arms, one training set. Bars, medians. \textbf{b}, Learned dose response, the ratio of the AUCs predicted for the same patient at 850 and 350~mg, against the exposure ratio between the training arms (one point per training set). Dashed, the arm ratio attributed to the dose. Dotted, the true ratio. \textbf{c}, Median signed AUC error at 350~mg for patients observed on 800~mg (bars, means).}
\label{fig:ebvae}
\end{figure}

In clinical data, the dose depends on the patient, because therapeutic drug monitoring adjusts it towards a target exposure~\cite{ahn2005inherent}. A model may then learn which patients receive which dose and not what the dose does. We tested this on the design of Baaz et al.\ with sparse sampling, by giving the high dose preferentially to fast eliminators (titrated arms), so that the two training arms ended with similar exposures (Fig.~\ref{fig:ebvae}a). The error of the empirical Bayes model on the five regimens of Baaz et al.\ more than doubled, from 14.4\% to 32.4\%, while its fit to the training regimen barely changed. The error of the dose-event model rose from 10.9\% to 12.3\%, and it was lower than that of the empirical Bayes model in each of 10 training runs (Table~\ref{tab:regimens}a).

To understand this difference, we measured what each model learned as the effect of the dose on one patient, the ratio of the AUCs it predicts for the same test patient under 850 and 350~mg. The ratio of the empirical Bayes model followed the exposure ratio between its two training arms, as expected for a model that attributes this ratio to the dose, and the ratio of the dose-event model stayed close to the true one (Fig.~\ref{fig:ebvae}b). With titrated arms, the dose response of the empirical Bayes model fell to about half of the true one (0.46, against 0.81 for the dose-event model). Patients observed on 800~mg and predicted at 350~mg were over-predicted by a median of 88\%, against 8\% with the dose-event model (Fig.~\ref{fig:ebvae}c). The same flattening appeared with randomised arms, in a training set whose 800~mg arm had by chance received faster eliminators (Supplementary Note~3).

We then tested where the empirical Bayes model attributes the difference between the arms to the dose (Supplementary Fig.~3). Removing the dose from its prior did not change its dose response (Supplementary Note~2). The dose-event model, in turn, lost its dose response in the same way once its encoder could also read the dose (Supplementary Table~2). Titration also affected the dose-shift and dose-in-the-dynamics models differently when training patients were few. Trained on doses titrated on the central volume, the dose-in-the-dynamics model attributed 37\% of the variance of its learned dynamics to the dose, against 67\% when trained on random doses. With 200 training patients, it appeared to infer the central volume from the administered dose, and the dose-shift model showed little evidence of this behaviour. This difference was no longer observed with 800 training patients (Supplementary Note~3).

\subsection*{Adapted from simulation, the models forecast the next monitoring visit in kidney transplant recipients}

\begin{table}[tb]
\centering
\footnotesize
\setlength{\tabcolsep}{3pt}
\caption{\textbf{Prediction of the next monitoring visit in kidney transplant recipients.} 91 pairs of consecutive visits in 32 patients, 51 with a dose change of at least 20\%. Every approach receives the samples at 0, 1 and 3~h of visit~$k$ and the dose of visit $k+1$. Median absolute percentage error of the AUC(0--12~h), mean (s.d.) over training runs. $k+1$, dose change, visit $k+1$ of these 51 pairs. Fine-tuned, on the pairs of the other patients, each patient held out in turn. $^{a}$From the full profile of visit~$k$. Details in Supplementary Table~6.}
\label{tab:limoges}
\begin{tabular}{@{}lccc@{}}
\toprule
\textbf{Approach} & \textbf{Visit $k$} & \textbf{Visit $k+1$} & \textbf{$k+1$, dose change} \\
\midrule
MAP-BE, published model & 5.2 & 24.5 & 25.5 \\
Dose-proportional rule$^{a}$ & -- & 24.8 & 26.4 \\
Unchanged AUC$^{a}$ & -- & 23.4 & 23.4 \\
\midrule
Dose-shift model, simulation only (10 runs) & 13.6 (5.9) & 32.6 (2.4) & 38.1 (4.7) \\
Dose-event model, simulation only (3 runs) & 18.4 (6.1) & 32.0 (1.4) & 35.4 (3.0) \\
Dose-shift model, fine-tuned (3 runs) & 8.8 (1.0) & \textbf{21.1} (0.4) & \textbf{22.4} (1.0) \\
Dose-event model, fine-tuned (3 runs) & 8.8 (0.5) & 21.6 (1.7) & 24.4 (2.2) \\
\bottomrule
\end{tabular}
\end{table}

No real patient is observed under two doses in the same state, so the closest available test of a dosing model is the next monitoring visit, at which the dose has often changed. We used 32 kidney transplant recipients with rich profiles at several visits, which gave 91 pairs of consecutive visits, 51 of them with a dose change of at least 20\%. From the samples at 0, 1 and 3~h of one visit and the dose of the next, each approach predicted the AUC of the next visit (Table~\ref{tab:limoges}).

Trained on simulated patients only, the neural models predicted the next visit with a median error of 32 to 33\%. Fine-tuning on the pairs of the other patients, each patient being held out in turn, reduced the error of the dose-shift model to 21.1\% and that of the dose-event model to 21.6\%. Bayesian estimation under the published population model reached 24.5\%, the dose-proportional rule applied to the full profile 24.8\% and an unchanged AUC 23.4\%. On the pairs with a dose change, the dose-shift model also had the lowest error (22.4\%), and its median signed error was $-5.1\%$, against $-21.0\%$ for Bayesian estimation (Supplementary Table~6). Its median error was lower than that of Bayesian estimation for 24 of the 32 patients. Adapted using the other 31 patients, a model pretrained on simulated patients forecast the exposure of each held-out patient with a lower median error than the approaches used in clinical practice. Although the neural models had lower median errors, the differences were modest and uncertain given the small number of patients.

\section*{Discussion}

Accuracy at the observed dose does not show that a model can support a change of dose. A neural model must also distinguish the effect of the dose from the characteristics of the patients who receive it, and how the dose enters the model matters because clinical doses are not randomly assigned. Population pharmacokinetic models can answer dosing questions because of their structure, but they can fail when their equations are wrong. The two models proposed here give this structure to models whose equations are learned. In both, the dynamics are generated from the state of the patient, and the dose moves the state and not the dynamics. The dose-shift model addresses a change of dose on the same schedule, and the dose-event model, which also keeps the drug separate from the physiology of the patient, addresses full regimens. The training data matter as well, since the largest gain in accuracy came from training on patients seen at two doses.

Without a predefined pharmacokinetic model to misspecify, the latent ODE predicted the AUC of the sampled tacrolimus interval with errors close to those of Bayesian estimation, and in kidney transplant recipients it was at least as accurate as the estimator used in clinical practice. The dosing decision, however, concerns an alternative dose, and there a wrong structure costs more. With linear instead of saturable elimination, Bayesian estimation fitted the observed visit almost as well, but its error at the alternative dose was 30\%, against 13\% for the dose-shift model, which needs no equations. A similar effect has been reported for voriconazole, where models without nonlinear clearance led to biased model-based doses~\cite{mcdougall2016impact}. Given the right equations, Bayesian estimation remained the most accurate, and a modeller comparing structures on patients seen at two doses can find them. A learned model may therefore be most useful when such a comparison cannot be made, for example when a published model is applied without new data at several doses. The saturable elimination of our simulator is also an assumption, because titration alone can create apparent nonlinearity in TDM data~\cite{ahn2005inherent}, so our comparison shows what a wrong structure can cost and does not estimate this cost for tacrolimus in patients.

Letting the dose move the latent state was about one point more accurate than giving the dose to the dynamics. Both designs can represent the same predictions at alternative doses, but they learn differently from the data that clinicians collect. Dynamics that depend on the dose learn their version at each dose only from the patients who received that dose~\cite{hernan2020causal}, and under TDM these patients are not comparable across doses. Dynamics that do not depend on the dose can be learned from patients receiving different doses. This reduces the risk of learning associations between prescribed doses and patient characteristics instead of the effect of the dose itself~\cite{ahn2005inherent}.

The dose-event model extends the same principle to a full dosing history. As in pharmacokinetic models, a dose is an addition to the drug state, linear in the amount, that leaves the physiology unchanged. This kept the model accurate on regimens other than the one it was trained on, including regimens fixed before the final runs, and under titration its dose response stayed at 0.81 of the true one while the error of the empirical Bayes model more than doubled. It is less accurate for a single change of dose, for which the dose-shift model remains the better choice.

Earlier neural models fall short for related reasons. The two-port model of Lu et al.\ places the dose correctly, but when it is trained on single visits it can leave the effect of the dose negligible~\cite{lu2021nmi}. Their regimen model adds each dose to a latent state that also describes the patient~\cite{lu2021neural}, so that an alternative regimen can also change what the model represents of the patient. In the empirical Bayes model~\cite{baaz2026empirical}, the dose enters the prior, the encoder and the learned dynamics, the last as in neural controlled differential equations~\cite{kidger2020neural}. With one regimen per patient, the difference between two training arms can come from the dose or from the patients who received it, and these routes let the model attribute it to the dose. In the dose-event model, the dose acts directly on the drug state, while patient characteristics are inferred from the observed concentrations. This helps prevent the model from confusing the prescribed dose with the characteristics of patients who receive it.

These results have limits. Accuracy at an alternative dose could only be measured in simulation, because no patient is observed under two doses in the same state~\cite{holland1986statistics}. On the two-visit cohort, alternative doses stayed within the training range. The regimen designs also included a loading dose above the training amounts, but this does not establish reliable predictions for arbitrary doses beyond the training range. Neither neural model learned saturation from patients observed under a single regimen, which both attributed to the patients on the high dose (Supplementary Note~3). Learning it requires data in which the same patient is seen under different doses, as in the two-visit design. The stability of the dose-event model under titration was shown on a linear simulator, and with six samples on the tacrolimus simulator it had no advantage over the empirical Bayes model. In real patients, the next monitoring visit is the closest test. There, the fine-tuned models had lower median errors than Bayesian estimation and the dose-proportional rule, so the approach can be adapted from simulated to real patients with little data. This result has to be weighed against the small number of patients and the modest differences, and an unchanged AUC came close (Supplementary Note~3). Between consecutive visits, haematocrit also rose and corticosteroid doses fell, so that the effect of the dose is mixed with physiological change. Separating the two requires a model in which they act separately, the dose through the drug coordinates and physiological change through a transition of the physiology coordinates, which the dose-event model makes possible.

In TDM software, where Bayesian estimators are used today (Fig.~\ref{fig:task}a), the dose-shift model would have predicted the AUC at an alternative dose within the $\pm20\%$ accepted for tacrolimus for 67.5\% of simulated patients, against 32.9\% for MAP-BE with linear elimination (Table~\ref{tab:benchmark}). Its prediction uncertainty was poorly calibrated in simulation (Supplementary Note~1). Before such a model is used to choose a dose, it has to be trained on the population and dose range where it will be used, and its predictive intervals have to be recalibrated. Its accuracy at alternative doses has to be measured on patients monitored under several doses, and its value has to be established prospectively. Nothing in this structure is specific to tacrolimus, and it could apply to other drugs whose dose is individualised from a few concentrations. The physiology coordinates also summarise the patient independently of the dose, which makes them a natural input for outcomes such as adverse effects or graft loss~\cite{tsiatis2004joint}, with the dose acting only through the drug state. Such extensions will require outcome data in which doses vary within patients.

\section*{Methods}

\subsection*{Study overview}
From a few samples collected at the current dose, the models predict the AUC at the observed dose, at an alternative dose on the same schedule, and under regimens with other dose times and amounts. Accuracy at an alternative dose can only be measured in simulation, where the exposure of every patient under any dose is known. Real patients test the observed dose and the next monitoring visit. Table~\ref{tab:data} gives the data of each experiment. Exact values and technical details are in Supplementary Note~4.

\subsection*{Clinical cohorts and ethical approval}
Data were collected from adult kidney transplant recipients who participated in a pharmacokinetic trial that complied with legal requirements and the Declaration of Helsinki. The protocol (NCT01435291, registered on 15 September 2011) was approved by the regional ethics committee (CPP Sud Méditerranée V) and all patients gave written informed consent. The development cohort (137 profiles from 32 patients) and the external cohort (72 patients) are described in refs.~\cite{woillard2011population,marquet2018comparative}. Profiles have 10--12 samples. The model input is the subset at 0, 1 and 3~h, and the reference AUC is computed from the full profile. Repeated cross-validation used 46 random splits of the profiles, not of the patients, so almost every test profile had other visits of the same patient in the training set.

The multi-visit cohort comprises 50 kidney transplant recipients with rich profiles at up to five visits. A profile was kept if it had a pre-dose sample, samples close to 1 and 3~h and at least six samples. Consecutive kept profiles of the same patient with the same dosing interval form a pair (91 pairs from 32 patients, all on twice-daily tacrolimus). The reference AUC(0--12~h) is the trapezoidal area of the observed profile, with the concentration at 12~h set to the pre-dose concentration when the profile ended earlier (72 pairs).

\begin{table}[tbp]
\centering
\footnotesize
\setlength{\tabcolsep}{2pt}
\caption{\textbf{Data used for each experiment.} Section, of the Results. Samples, number of samples per visit given to the model. CV, cross-validation. Each training patient of the two-visit cohorts gives two training pairs (visit 1 to visit 2, visit 2 to visit 1). $^{a}$Visit-1 average concentration restricted to 3--25~ng\,mL$^{-1}$.}
\label{tab:data}
\begin{tabular}{@{}lllll@{}}
\toprule
\textbf{Section} & \textbf{Data} & \textbf{Training patients} & \textbf{Test patients} & \textbf{Samples} \\
\midrule
Observed dose & Scenarios 1--3, simulated & 160 & 40, 100 datasets & 3 \\
 & Development cohort, real & 137 profiles, 32 patients & 46 CV splits & 3 \\
 & External cohort, real & development cohort & 72 & 3 \\
\midrule
Alternative dose & Two-visit cohort$^{a}$ & 100 & 1100 & 3 \\
 & Two-visit cohort, variants & 100 & 1000 & 3 \\
 & Dose-switch cohort & 100 & 1261 & 3 \\
\midrule
Unseen regimens & Design of Baaz et al. & 200 & 400 & 6 or 61 \\
 & Tacrolimus design & 200 & 400 & 6, 12 or 61 \\
\midrule
Adjusted doses & Design of Baaz et al., titrated & 200 & 400 & 6 \\
 & Paired cohorts, Michaelis constant & 800 & 1000 & 3 \\
 & Paired cohorts, central volume$^{a}$ & 200, 400 or 800 & 1410 and 1058 & 3 \\
\midrule
Next visit & 91 pairs of visits, real & other patients & 32 patients & 3 \\
\bottomrule
\end{tabular}
\end{table}

\subsection*{Simulated patients}
The tacrolimus simulator is a three-transit absorption, two-compartment model with Michaelis--Menten elimination, with typical values from the published model~\cite{woillard2011population} and log-normal between-subject variability. In the two-visit cohort, each patient has two visits at doses drawn independently of physiology between 1 and 8~mg. A patient was retained only if the true average concentration over visit~1 lay within 3--25~ng\,mL$^{-1}$, which correlates the visit-1 dose with physiology, as titration does in practice. In the dose-switch cohort, each patient is at steady state on a first dose, is switched to a second and is observed four days later. The scenarios of Table~\ref{tab:scenarios} use the same absorption and disposition, with linear or saturable elimination.

The regimen design of Baaz et al.~\cite{baaz2026empirical} has one compartment and linear elimination, and we checked our implementation against their published function and data. Training patients receive 400 or 800~mg, with 6 or 61 samples. Arms are randomised or titrated, with the high dose given preferentially to fast eliminators. New patients are encoded from their samples under the training regimen and predicted under the regimens of Baaz et al.\ (new amounts and changes of schedule) and, for the frozen model, under five further regimens, including a loading, a missed and an additional dose. The tacrolimus regimen design transposes this design to the tacrolimus simulator, with linear or saturable elimination. The error is the median absolute relative AUC error against the noise-free curve, averaged over the regimens.

\subsection*{Models}
\paragraph{Plain latent ODE.} An ODE-RNN encoder~\cite{rubanova2019latent} reads the samples of a visit with their dose and the static covariates, and outputs a Gaussian posterior over a latent state, with a Gaussian mixture prior~\cite{dilokthanakul2016deep}. A neural network defines the dynamics $\dot z=f(z)$, without a dose input, integrated with an adaptive Dormand--Prince solver~\cite{dormand1980family}. A readout returns the concentration, and the AUC is integrated from the predicted curve. This model is trained on single visits.

\begin{figure}[tbp]
\centering
\includegraphics[width=0.85\textwidth]{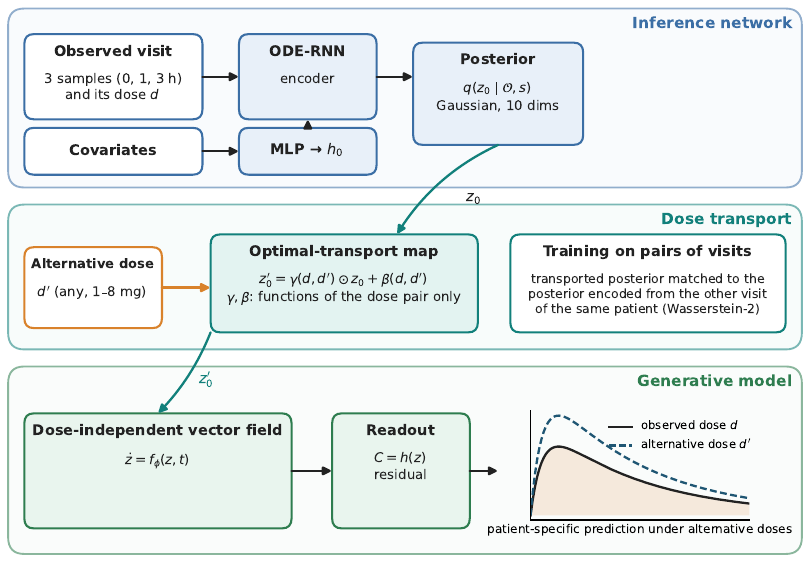}
\caption{\textbf{Architecture of the dose-shift model.} The encoder maps the samples of the observed visit, its dose and the covariates $s$ to a Gaussian posterior over the latent state. For an alternative dose $d'$, an affine map whose coefficients depend on the dose pair moves the encoded state. Dynamics that do not depend on the dose are integrated from the moved state, and a readout returns the concentration curve.}
\label{fig:otfilm}
\end{figure}

\paragraph{Dose-shift model (OT-FiLM).} The encoder, dynamics and readout are those of the plain latent ODE (Fig.~\ref{fig:otfilm}). Two networks of the dose pair $(d,d')$ move the initial state as $z_0'=\gamma\had z_0+\beta$ before integration, a feature-wise linear modulation~\cite{perez2018film}. This map has the form of the optimal transport between two diagonal Gaussian distributions~\cite{givens1984class}, hence the name OT-FiLM. We make no claim that the learned latent coordinates are identifiable or physiologically interpretable~\cite{locatello2019challenging}. The model is trained on pairs of visits of the same patient, in both directions. From the samples of one visit, it reconstructs the profiles of both visits, in an importance-weighted bound~\cite{burda2015importance}. One penalty brings the moved samples towards the posterior encoded from the other visit, and another makes an unchanged dose leave the state unchanged. The last checkpoint is evaluated, never one selected on test data.

\begin{figure}[tbp]
\centering
\includegraphics[width=0.85\textwidth]{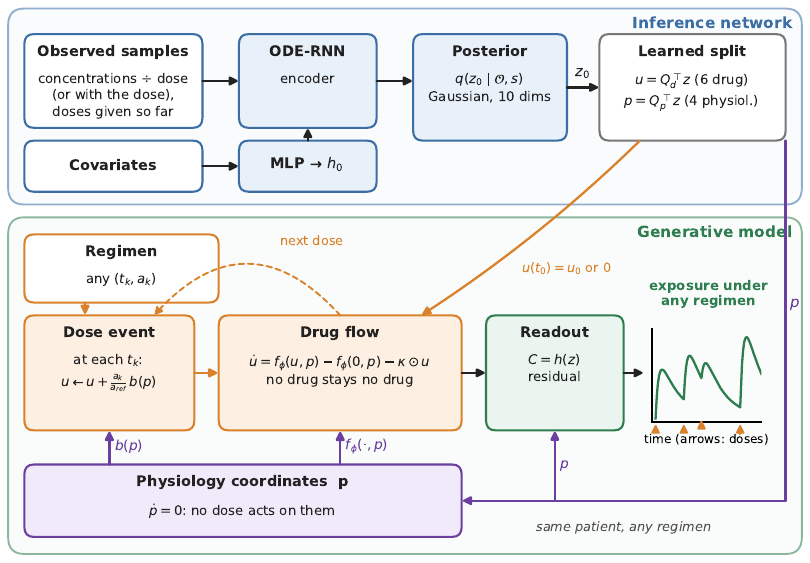}
\caption{\textbf{Architecture of the dose-event model.} The encoder maps the observed samples and the covariates $s$ to a Gaussian posterior over the latent state, which a learned orthonormal basis $Q$ splits into drug coordinates $u$ and physiology coordinates $p$. Each administration $(t_k,a_k)$ translates $u$ by $a_k\,b(p)$. Between administrations, a drug network moves only $u$, and the readout uses $u$ and $p$.}
\label{fig:arch-de}
\end{figure}

\paragraph{Dose-event model.} A learned orthonormal basis splits the latent state into drug coordinates $u$ and physiology coordinates $p$ (Fig.~\ref{fig:arch-de}). An administration of amount $a$ translates $u$ by $a\,b(p)$, linear in the amount because simultaneous administrations add. Between administrations, $\dot u=f(u,p)-f(0,p)-\kappa\odot u$, so a patient without drug stays without drug and no compartmental structure is imposed. The flow acts only on $u$, so $p$ is constant along every trajectory and an alternative regimen can be simulated with the physiology of the patient unchanged. Every nonlinearity in how a dose becomes a concentration is left to the flow, since an additive input into a dynamical system followed by a nonlinear readout can approximate any time-invariant dose--concentration map with fading memory~\cite{boyd1985fading}. On the two-visit cohort, each visit is generated from the drug-free state through its dosing history. On the regimen designs, patients start with drug on board, and the encoder reads the concentrations divided by the dose of the patient's arm, but not the dose.

\subsection*{Comparators}
The dose-in-the-dynamics model is the dose-shift model without the dose map, with the dose given to its dynamics and readout, and it is trained with the evidence lower bound and not the importance-weighted bound. The dose-input model is the dose-event model without the latent split, with a decaying trace of the doses as an input of its dynamics. It was tuned on the training-regimen error before any comparison.

We implemented the two-port model of Lu et al.~\cite{lu2021nmi} in its original configuration and with the covariates and the cross-visit training added, with the default settings of our implementation. Their regimen model~\cite{lu2021neural} comes from their public code and is used as that code predicts an alternative regimen. On the design of Baaz et al., both models were run on development training sets only, not on the five further regimens. The empirical Bayes model uses the public code of ref.~\cite{baaz2026empirical}, unmodified on its own design.

For Bayesian estimation on the two-visit cohort, population models were fitted by SAEM (Monolix) on the training patients, with saturable elimination, the structure of the simulator, or with linear elimination. The two structures were also fitted with the same covariates and compared by the Bayesian information criterion. Individual parameters were estimated by maximum a posteriori from the samples given to the other models. The dose-proportional rule multiplies the estimated AUC by the ratio of the doses. On the design of Baaz et al., the population model has the structure of the simulator.

\subsection*{Prediction at the observed dose}
The population model of Table~\ref{tab:scenarios} has linear elimination and no haematocrit effect, and individual parameters are estimated from the three test samples. The latent ODE is trained on the same patients with a linear readout, and evaluated at the checkpoint with the lowest smoothed reconstruction error on the test patients during training, which favours the latent ODE. Table~\ref{tab:scenarios} reports the untrimmed RMSPE over the 100 datasets. The mean and standard deviation of the latent ODE in Scenario~3 come from our rerun of the pipeline on 100 datasets, because the standard deviation of our earlier analysis, 0.04, could not be reproduced. The other values of Table~\ref{tab:scenarios} come from that earlier analysis. For the plain latent ODE, normalisation constants (the maximum concentration and the Box--Cox parameter) come from the training and test patients. In the datasets of our rerun, the maximum was unchanged without the test patients in 82\% of datasets and within a factor of 1.1 in 90\% (1.82 at most), and the Box--Cox parameter changed by at most 0.05.

\subsection*{Prediction at an alternative dose}
For the dose-shift and dose-in-the-dynamics models, normalisation constants are those of the training cohort. The variants of Supplementary Table~1 use a cohort without the visit-1 exposure restriction.

\subsection*{Prediction under unseen regimens}
The dose-event model was chosen among the variants of Supplementary Table~2, on development runs and on the same five regimens of Baaz et al. The runs of Table~\ref{tab:regimens}a were made afterwards, with the model frozen and the comparisons specified in advance, on new training sets and new test patients. One run of the empirical Bayes model diverged during training and is left out with its pair, a rule set after the runs. Population-level predictions, the metric reported by Baaz et al., are not compared, since the dose-event model has no learned population prior. In the tacrolimus design, predicted concentrations below zero were set to zero for every model without a positive readout. The patient of Fig.~\ref{fig:tacro-patients} was chosen by a rule fixed before plotting, as the test patient of the 2-mg arm with the median true AUC.

\subsection*{Doses adjusted to the patient}
The dose response of a model is the ratio of the AUCs it predicts for the same patient under 850 and 350~mg, divided by the true ratio. In the confounding experiments, paired cohorts contain the same patients and differ only in the visit-1 dose, which is correlated with the confounder ($\Km$ or $\Vc$) in the confounded cohort and not in the control cohort. The endpoint, fixed in advance, is a difference-in-differences of the median AUC error at visit~2. This is the extra error of the confounded-trained model over the control-trained model where the correlation is removed, minus the same quantity where the correlation still holds. The difference removes the cost of reduced coverage that confounded training imposes on both designs.

\subsection*{Next monitoring visit}
Every approach received the samples at 0, 1 and 3~h of the first visit of a pair, the covariates of that visit and the dose of the second. MAP-BE used the published population model~\cite{woillard2011population}. The dose-proportional rule multiplied the reference AUC of the first visit by the dose ratio, and the unchanged-AUC rule took that reference AUC as the prediction. Fine-tuning held out each patient in turn and continued training from the simulation-trained model on the pairs of the other patients, with all settings fixed before any run. Every approach was scored on the observed samples of the held-out patient.

\subsection*{Statistics and reproducibility}
For each patient, the relative AUC error is $\hat A/A-1$. We report its median absolute value, the median signed value (bias), the share within $\pm20\%$ and the count above 100\%. RMSPE is used only in Table~\ref{tab:scenarios} and for the clinical cohorts at the observed dose, because a few patients with small true AUCs dominate it (in one evaluation, a single patient of 1000 contributed 78\% of the squared error). Networks are trained for a fixed budget, evaluated at their last iteration and summarised by the mean and standard deviation over training runs, except in Table~\ref{tab:scenarios}. For paired comparisons we averaged each patient's absolute error over runs, took the difference of the medians, and computed a 95\% interval by resampling patients (2000 bootstrap replicates) and a Wilcoxon signed-rank test. These intervals reflect patient sampling, not training variability. The comparisons of Table~\ref{tab:regimens}a and the endpoint of the confounding experiments were fixed in advance.

\subsection*{Language model usage}
Large language models (ChatGPT, Claude) were used for copy-editing and for assistance with analysis code. All analyses were checked by the authors.

\section*{Data availability}
The clinical datasets are not publicly available because of patient privacy regulations (GDPR). Anonymised data are available from J.-B.W.\ on reasonable request, subject to a data-sharing agreement and approval by the ethics committee. The simulated cohorts can be regenerated with the code at \url{https://github.com/BenJMaurel/structured-neural-odes}.

\section*{Code availability}
Code for the models, simulators, baselines and the tables and figures from simulated patients is available under the MIT licence at \url{https://github.com/BenJMaurel/structured-neural-odes}. The code that reads the clinical data is not included.

\section*{Acknowledgements}
This work was supported by the DIGPHAT project, funded by the French government and managed by the National Research Agency (ANR) under the France 2030 programme (ANR-22-PESN-0017).

\section*{Author contributions}
B.M.\ conceptualised the methodology, developed the software, performed the analyses and wrote the original draft. A.G.\ contributed to the methodology and reviewed the manuscript. S.Z.\ reviewed and edited the manuscript. M.U.\ and J.-B.W.\ acquired funding, supervised the research, contributed to the conceptualisation, and reviewed and edited the manuscript. All authors approved the final version.

\section*{Competing interests}
The authors declare no competing interests.

\clearpage
\includepdf[pages=-]{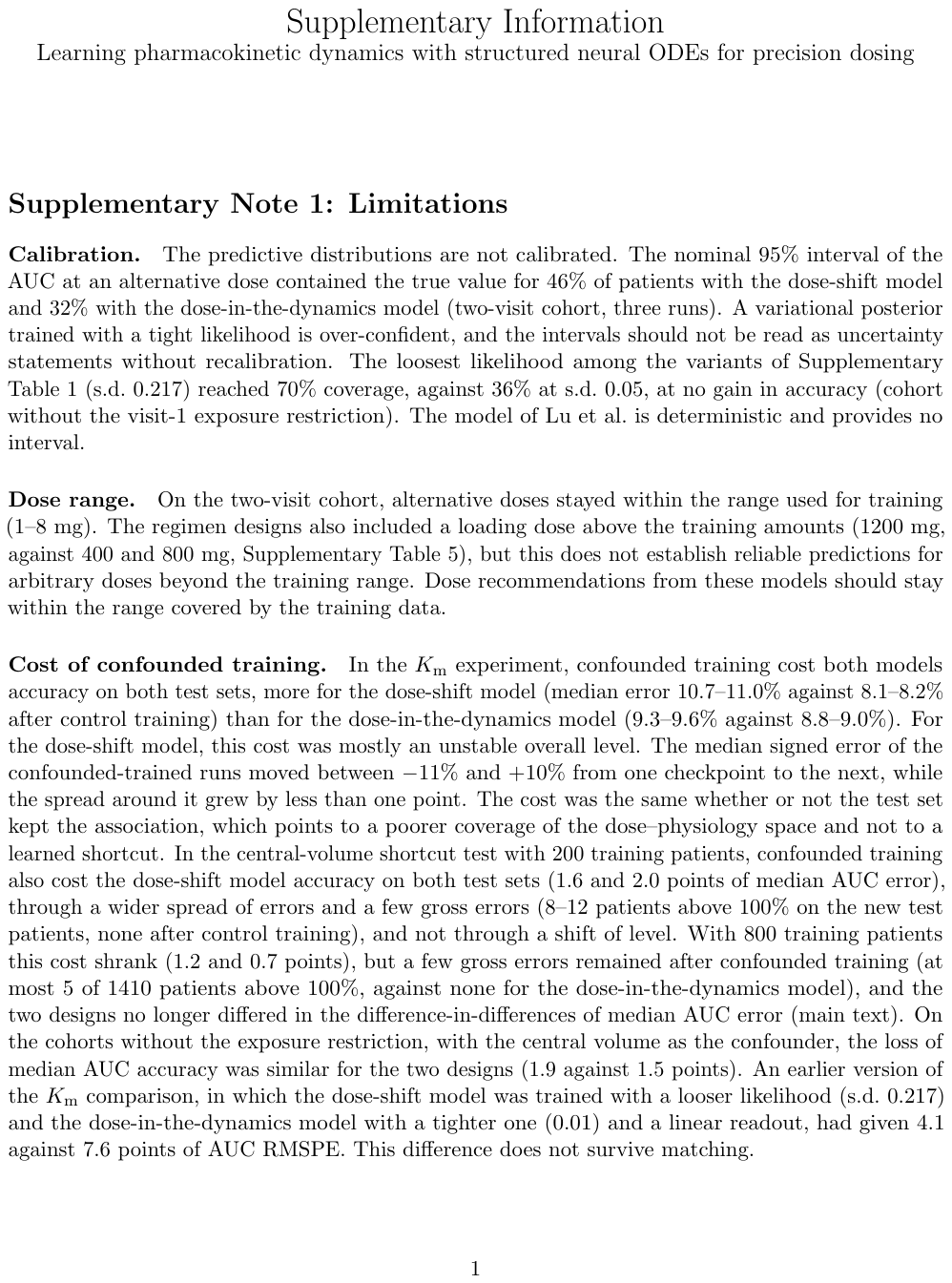}
\end{document}